\documentclass[runningheads,a4paper]{llncs}

\usepackage{graphicx}
\usepackage{subfigure}
\usepackage{url}
\usepackage[multiple]{footmisc}

\newcommand{\todo}[1]{{}}

\begin{document}

\title{Establishing linguistic conventions in task-oriented primeval dialogue}

\author{Martin Bachwerk \and Carl Vogel}

\institute{Computational Linguistics Group, \\
School of Computer Science and Statistics, \\
Trinity College, Dublin 2, Ireland \\
\email{\{bachwerm,vogel\}@tcd.ie}}

\maketitle

\begin{abstract}
In this paper, we claim that language is likely to have emerged as a mechanism for coordinating the solution of complex tasks. To confirm this thesis, computer simulations are performed based on the coordination task presented by Garrod \& Anderson (1987). The role of success in task-oriented dialogue is analytically evaluated with the help of performance measurements and a thorough lexical analysis of the emergent communication system. Simulation results confirm a strong effect of success mattering on both reliability and dispersion of linguistic conventions.
\end{abstract}

\section{Introduction} \label{int}

In the last decade, the field of communication science has seen a major increase in the number of research programmes that go beyond the more conventional studies of human dialogue (e.g. \cite{Garrod1987,Garrod1994}) in an attempt to reproduce the emergence of conventionalized communication systems in a laboratory (e.g. \cite{Galantucci2005,Garrod2007,Healey2002}). In his seminal paper, Galantucci has proposed to refer to this line of research as \textit{experimental semiotics}, which he sees as a more general form of \textit{experimental pragmatics}. In particular, Galantucci defines that the former \textit{``studies the emergence of new forms of communication''}, while the latter \textit{``studies the spontaneous use of pre-existing forms of communication''} (p. 394, \cite{Galantucci2009}).

Experimental semiotics provides a novel way of reproducing the emergence of a conventionalized communication system under laboratory conditions. However, the findings from this field cannot be transferred to the question of primeval emergence of language without the caveat that the subjects of the present-day experiments are very much familiar with the concepts of conventions and communication systems (even if they are not allowed to employ any existing versions of these in the conducted experiments), while our ancestors who somehow managed to invent the very first conventionalized signaling system, by definition, could not have been aware of these concepts. Since experimental semiotics researchers cannot adjust the minds of their subjects in order to find out how they could discover the concept of a communication system, the most these experiments can realistically achieve is make the subjects signal the `signalhood' of some novel form of communication (see. \cite{Scott-Phillips2009}). To go any further seems at least for now to require the use of computer models and simulations. 

Consequently, we are interested in how a community of simulated agents can agree on a set of lexical conventions with a very limited amount of given knowledge about the notion of a communication system. In this particular paper, we address this issue by conducting several computer simulations that are meant to reconstruct the human experiments conducted by \cite{Garrod1987} and \cite{Garrod1994}, which suggest that the establishment of new conventions requires for at least some understanding to be experienced, for example measured in the success of the action performed in response to an utterance, and that differently organized communities can come up with variously effective communication systems. While the communities in the current experiments are in a way similar to the social structures implemented in \cite{Bachwerk2010}, the focus here is on local coordination and the role of task-related communicative success, rather than the effect of different higher-order group structures.

\section{Modelling Approach}

The experiments presented in this paper have been performed with the help of the Language Evolution Workbench (LEW) (see \cite{Vogel2006,Bachwerk2010} for more detailed descriptions of the model). This workbench provides over 20 adjustable parameters and makes as few assumptions about the agents' cognitive skills and their awareness of the possibility of a conventionalized communication system as possible. The few cognitive skills that are assumed can be considered as widely accepted (see \cite{Jackendoff1999,Tomasello2003} among others) as the minimal prerequisites for the emergence of language. These skills include the ability to observe and individuate events, the ability to engage in a joint attention frame fixed on an occurring event, and the ability to interact by constructing words and utterances from abstract symbols\footnote{While we often refer to such symbols as `phonemes' throughout the paper, there is no reason why these should not be representative of gestural signs.} and transmitting these to one's interlocutor.\footnote{Phenomena such as noise and loss of data during signal transmission are ignored in our approach for the sake of simplicity.}\footnote{It is important to stress out that hearers are not assumed to know the word boundaries of an encountered utterance. However, simulations with so called synchronized transmission have been performed previously by \cite{Vogel2010}.} During such interactions, one of the agents is assigned the intention to comment on the event, while a second agent assumes that the topic of the utterance relates in some way to the event and attempts to decode the meaning of the encountered symbols accordingly.

From an evolutionary point of view, the LEW fits in with the so called faculty of language in the narrow sense as proposed by \cite{Hauser2002} in that the agents are equipped with the sensory, intentional and concept-mapping skills at the start, and the simulations attempt to provide an insight into how these could be combined to produce a communication system with comparable properties to a human language. From a pragmatics point of view, our approach directly adopts the claim made by \cite{Pickering2004} that dialogue is the underlying form of communication. Furthermore, despite the agents in the LEW lacking any kind of embodiment, they are designed in a way that makes each agent individuate events according to its own perspective, which in most cases results in their situation models being initially non-aligned, thus providing the agents with the task of aligning their representations, similarly to the account presented in \cite{Pickering2004}.

\section{Experiment Design}

In the presented experiments, we aim to reproduce the two studies originally performed by Garrod and his colleagues, but in an evolutionary simulation performed on an abstract model of communication. Our reconstruction lies in the context of a simulated dynamic system of agents which should provide us with some insights about how Garrod's findings can be transferred to the domain of language evolution. The remainder of this section outlines the configuration of the LEW used in the present study, together with an explanation of the three manipulated parameters. The results of the corresponding simulations are then evaluated in the following section \ref{res}, with special emphasis being put on the communicative potential and general linguistic properties of the emergent communication systems.\footnote{We intentionally refrain from referring to the syntax-less communication systems that emerge in our simulations as `language' as that would be seen as highly contentious by many readers. Furthermore, even though the term `protolanguage' appears to be quite suited for our needs (cf. \cite{Jackendoff1999}), the controversial nature of that term does not really encourage its use either, prompting us to stick to more neutral expressions.}

Garrod observed in his two studies that conventions have a better chance of getting established and reused if their utilisation appears to lead to one's interlocutor understanding of one's utterance, either by explicitly signaling so or by performing an adequate action. Notably, in task-based communication, interlocutors may succeed in achieving a task with or without complete mutual understanding of the surrounding dialogue. Nevertheless, our simulations have been focussed on a parameter of the LEW that defines the probability that \textit{communicative success matters} $p_{sm}$ in an interaction. From an evolutionary point of view, this parameter is motivated by the numerous theories that put cooperation and survival as the core function of communication (e.g. \cite{Bickerton2002}). However, the abstract implementation of the parameter allows us to refrain from selecting any particular evolutionary theory as the target one by generalizing over all kinds of possible success that may result from a communication bout, e.g. avoiding a predator, hunting down a prey or battling off a rival gang.

The levels of the parameter that defines if success matters were varied between 0 and 1 (in steps of 0.25) in the presented simulations. To clarify the selected values of the parameter, $p_{sm}$=$0$ means that communicative success plays no role whatsoever in the system and $p_{sm}$=$1$ means that only interactions satisfying a \textit{minimum success threshold} will be remembered by the agents. The minimum success threshold is established by an additional parameter of the LEW and can be generally interpreted as the minimum amount of information that needs to be extracted by the hearer from an encountered utterance in order to be of any use. In our experiments, we have varied between a minimum success threshold of 0.25 and 1 (in steps of 0.25).\footnote{Setting the minimum success threshold to 0 is equivalent to having $p_{sm}=0$.} The effects of this parameter will not be reported in this paper due to a lack of significance and space limitations.

In addition to the above two parameters, the presented experiments also introduce two different interlocutor arrangements, similar to the studies in \cite{Garrod1987} and \cite{Garrod1994}. In the first of these, pairs of agents are partnered with each other for the whole duration of the simulation, meaning that they do not converse with any other agents at all. The second arrangement emulates the community setting introduced in \cite{Garrod1994} by successively alternating the pairings of agents, in our case after every 100 interaction `epochs'.\footnote{In both cases, the agent population was set to ten and so each `epoch' comprised ten interactions, whereby every agent would on average take part in two interactions: once as a speaker and once as a hearer.} The introduction of the community setting was motivated by the hypothesis that a community of agents should be able to engage in a global coordination process, as opposed to local entrainment, resulting in more generalized and thus eventually more reliable conventions.

\section{Results and Discussion} \label{res}

The experimental setup described above resulted in 34 different parameter combinations, for each of which 600 independent runs have been performed in order to obtain empirically reliable data. The evaluation of the data has been performed with the help of a number of measures that have been selected with the goal of being able to describe both the communicative usefulness of an evolved convention system, as well as compare its main properties to those of languages as we know them now (see \cite{Bachwerk2010} for a more detailed account).

\begin{figure}[htb]
\centering
\subfigure{
\includegraphics[angle=-90,scale=0.40]{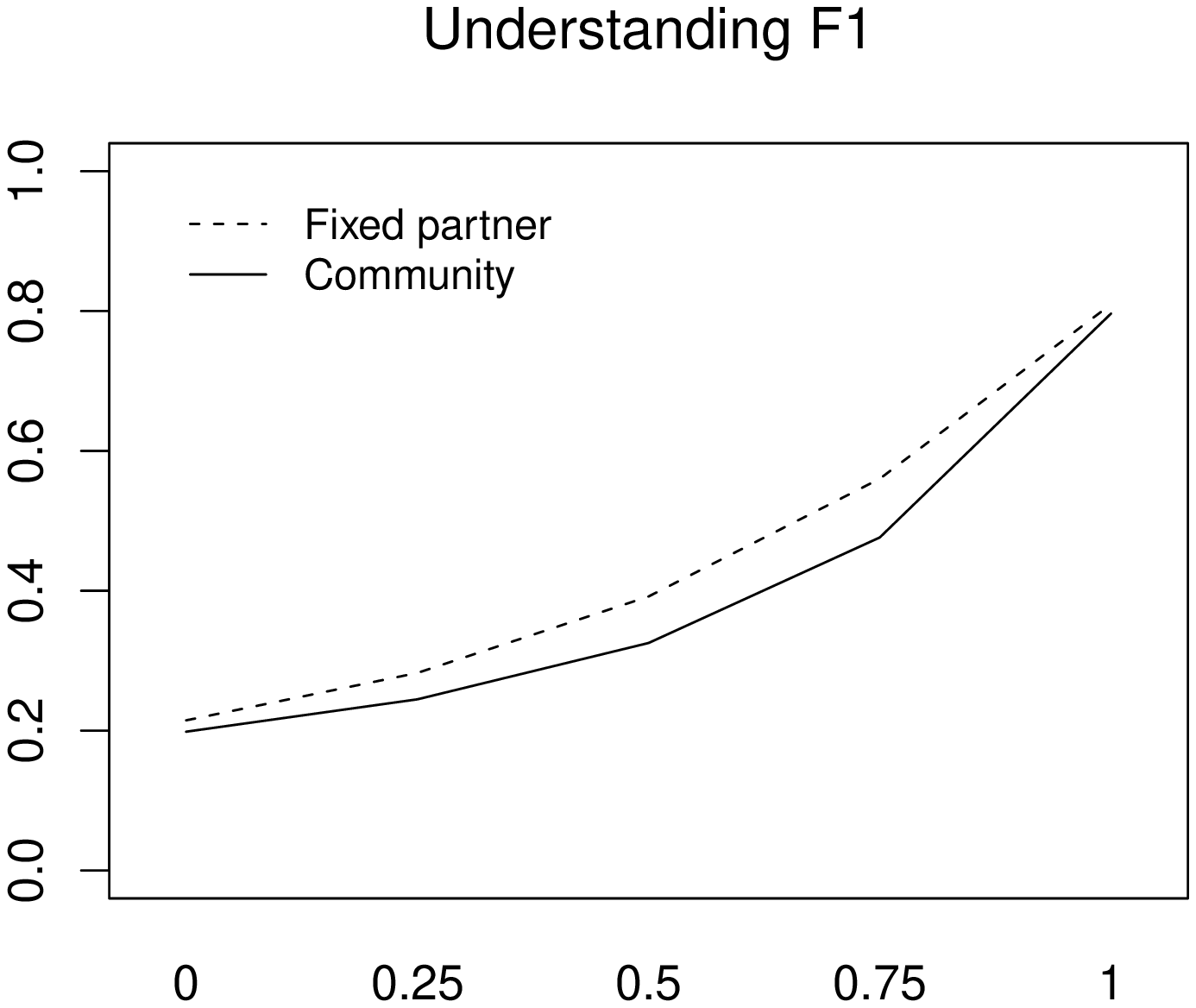}
\label{e1-f1}
}
\subfigure{
\includegraphics[angle=-90,scale=0.40]{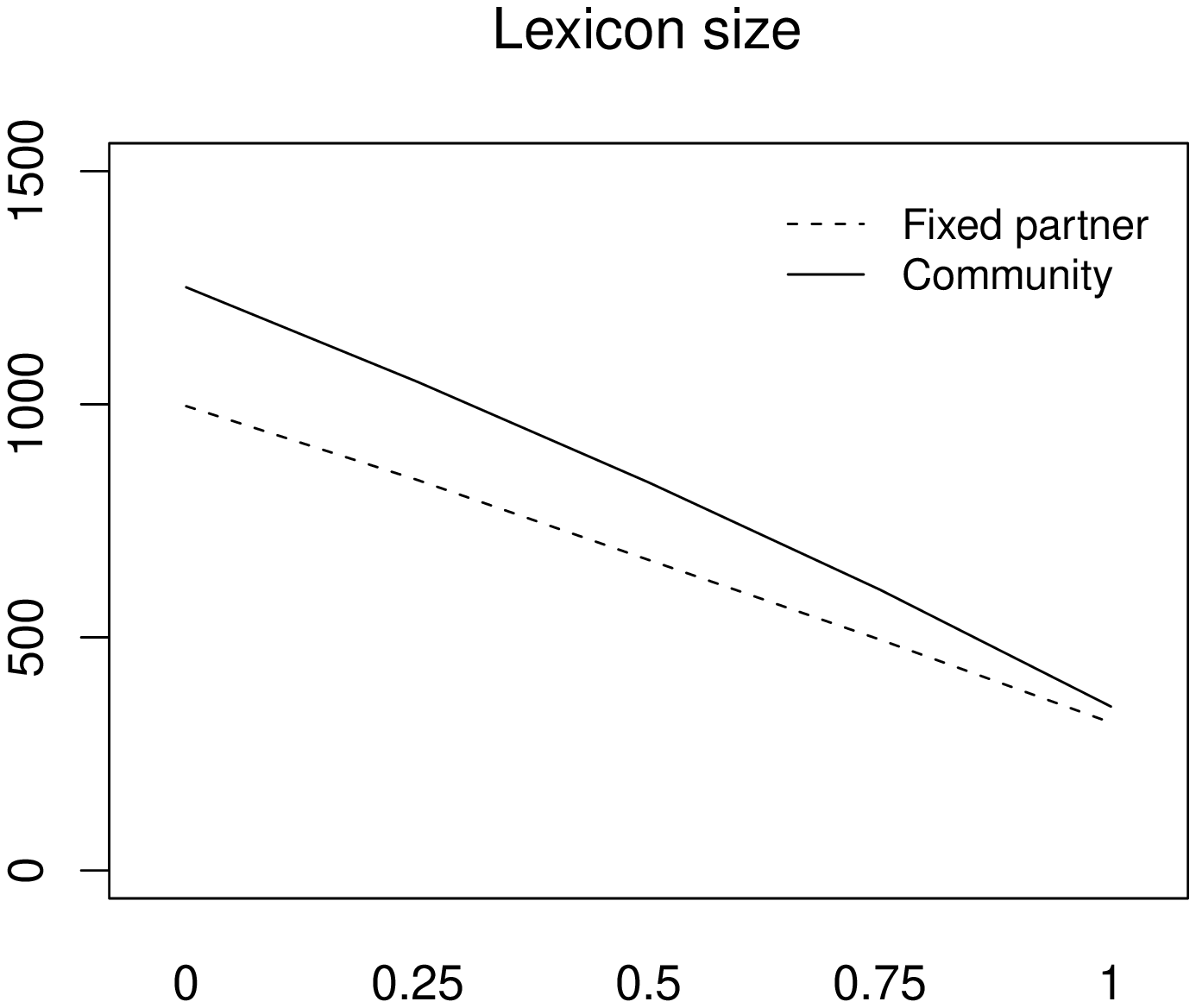}
\label{e1-lbsize}
}
\caption{{\footnotesize Effect of the interaction type and the probability that success matters on \subref{e1-f1} communicative success and \subref{e1-lbsize} agent lexicon size.}}
\end{figure}

In order to understand how well a communication system performs in a simulation, it is common to observe the understanding precision and recall rates, which can be combined to a single F-measure ($F1 = 2*\frac{precision*recall}{precision+recall}$). As can be seen from Figure \ref{e1-f1}, the results suggest that having a higher $p_{sm}$ has a direct effect on the understanding rates of a community ($t$ value between 26.68 and 210.63, $p<0.0001$). However, a communication setup in which agents communicate with each other in turns as opposed to with a fixed partner does not appear to be advantageous for the establishment of a reliable means of communication ($t=-15.85$, $p<0.0001$). Looking further, Figure \ref{e1-lbsize} indicates that, just as observed in \cite{Garrod1994}, agents operating in a community have a larger amount of variation available to them, in our case in the form of a larger lexicon ($t=35.52$, $p<0.0001$). However, unlike in the empirical study, the agents in the LEW do not benefit from this property, among other things due to the lack of an ability to enter into a negotiation about conventions to use in a given context.

It is important to note at this stage that the understanding measure presented in Figure \ref{e1-f1} only takes into account the interactions that have been successful, i.e. were not below the minimum success threshold in cases where success was chosen to matter. Consequently, this figure does not tell us how well the agents' lexicons are actually equipped to interpret a wide range of utterances. In order to evaluate the lexicons of agents without any effect that simple guessing luck might have on understanding, we take a look at two further measures: lexicon use, i.e. the average ratio of forms of an utterance that the hearer agent was able to find in his lexicon, and lexicon precision, i.e. the ratio of correct meanings found by the hearer, in the cases where the agent used his lexicon for decoding a form. Furthermore, the decrease in lexicon size alone does not provide any specific information as to what exactly is happening to the agents' lexicons. In other terms, further measures are required that could explain what effect the decrease actually has on the expressive and interpretative potential of a lexicon.

\begin{figure}[htb]
\centering
\subfigure{
\includegraphics[angle=-90,scale=0.40]{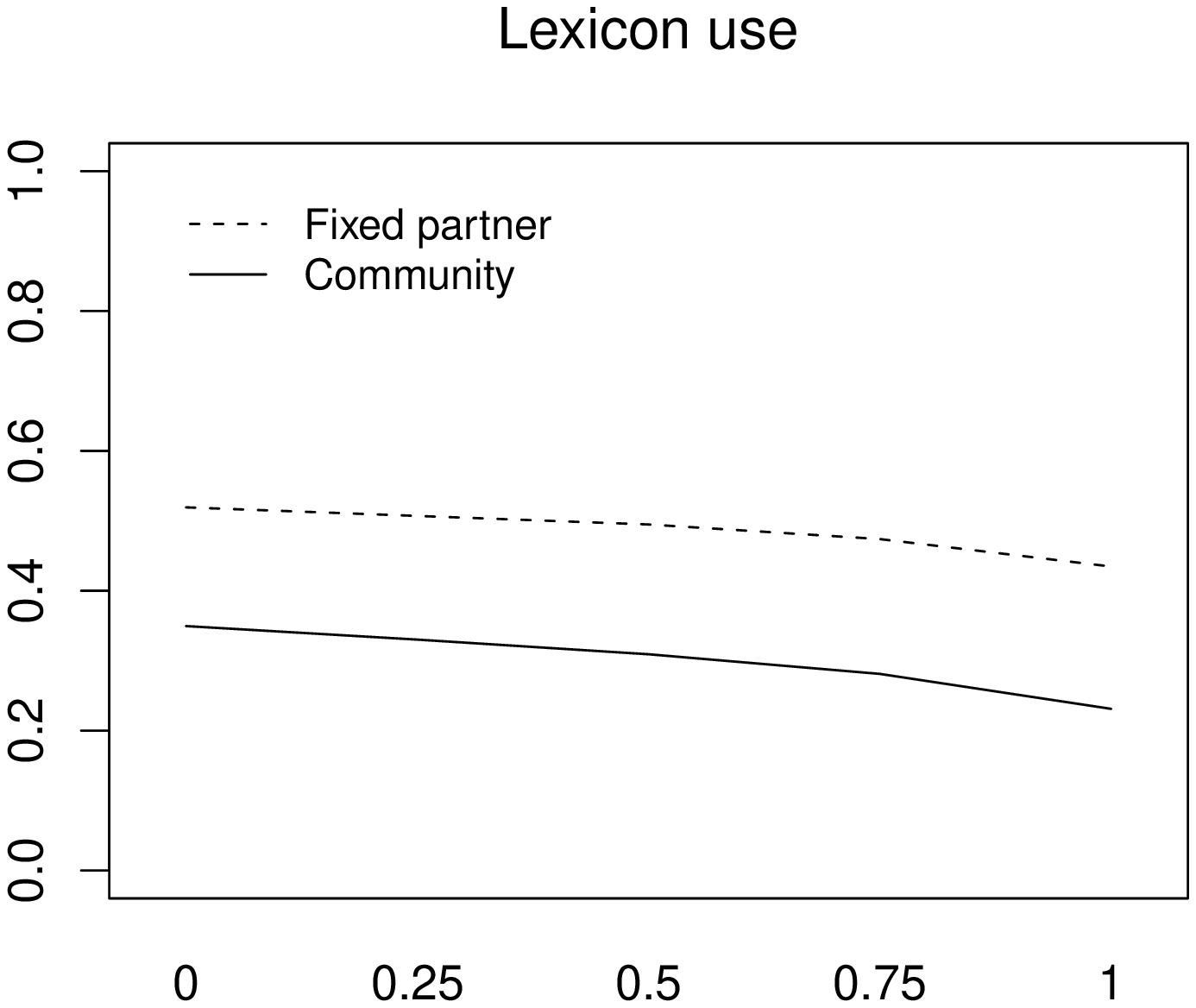}
\label{e1-lbuse}
}
\subfigure{
\includegraphics[angle=-90,scale=0.40]{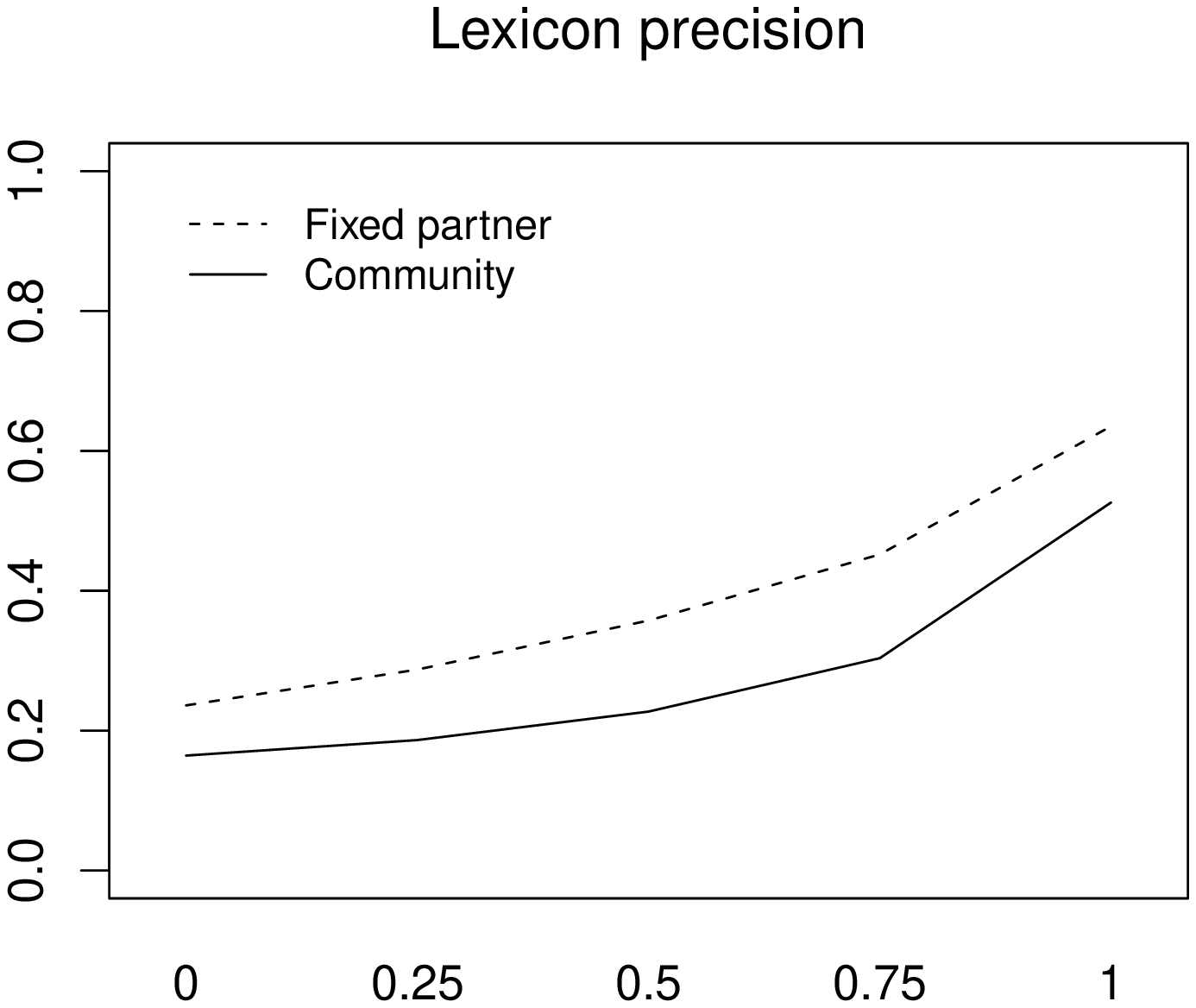}
\label{e1-lbprec}
}
\caption{{\footnotesize Effect of the interaction type and the probability that success matters on \subref{e1-lbuse} lexicon use and \subref{e1-lbprec} lexicon precision.}}
\end{figure}

\begin{figure}[htb]
\centering
\subfigure{
\includegraphics[angle=-90,scale=0.40]{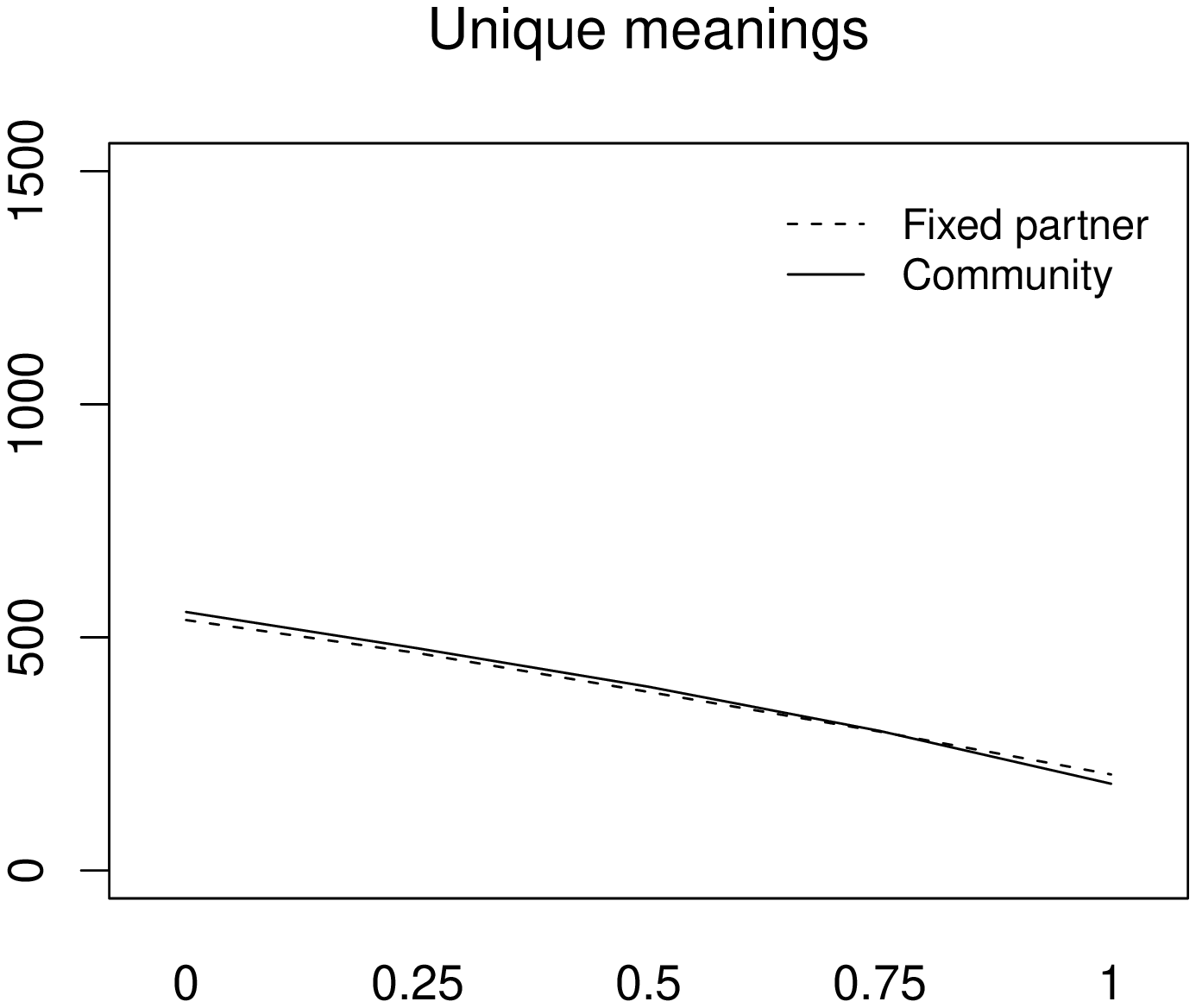}
\label{e1-m}
}
\subfigure{
\includegraphics[angle=-90,scale=0.40]{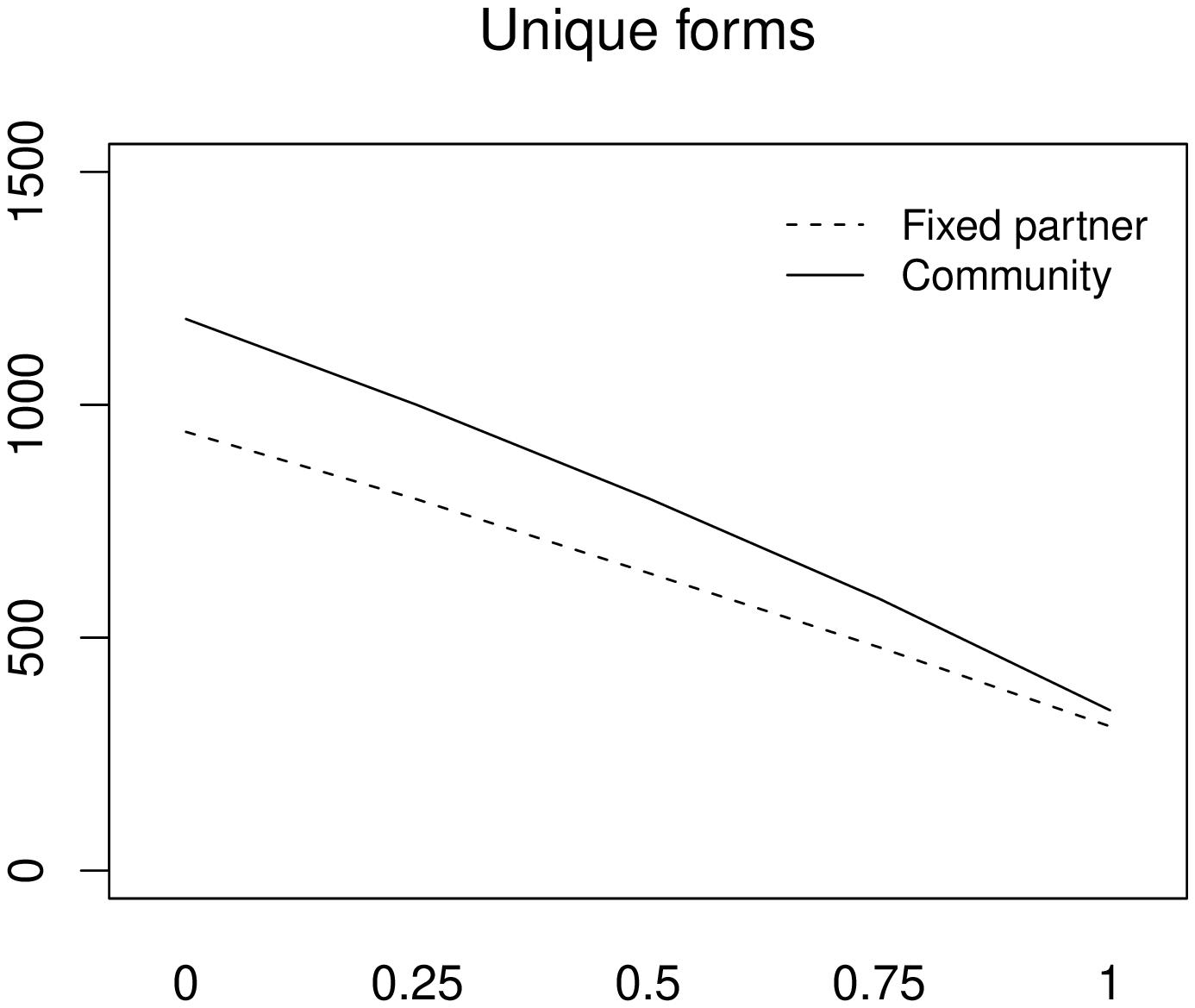}
\label{e1-f}
}
\caption{{\footnotesize Effect of the interaction type and the probability that success matters on the number of \subref{e1-m} unique meanings and \subref{e1-f} unique forms in agent lexicons.}}
\end{figure}

\begin{figure}[htb]
\centering
\subfigure{
\includegraphics[angle=-90,scale=0.40]{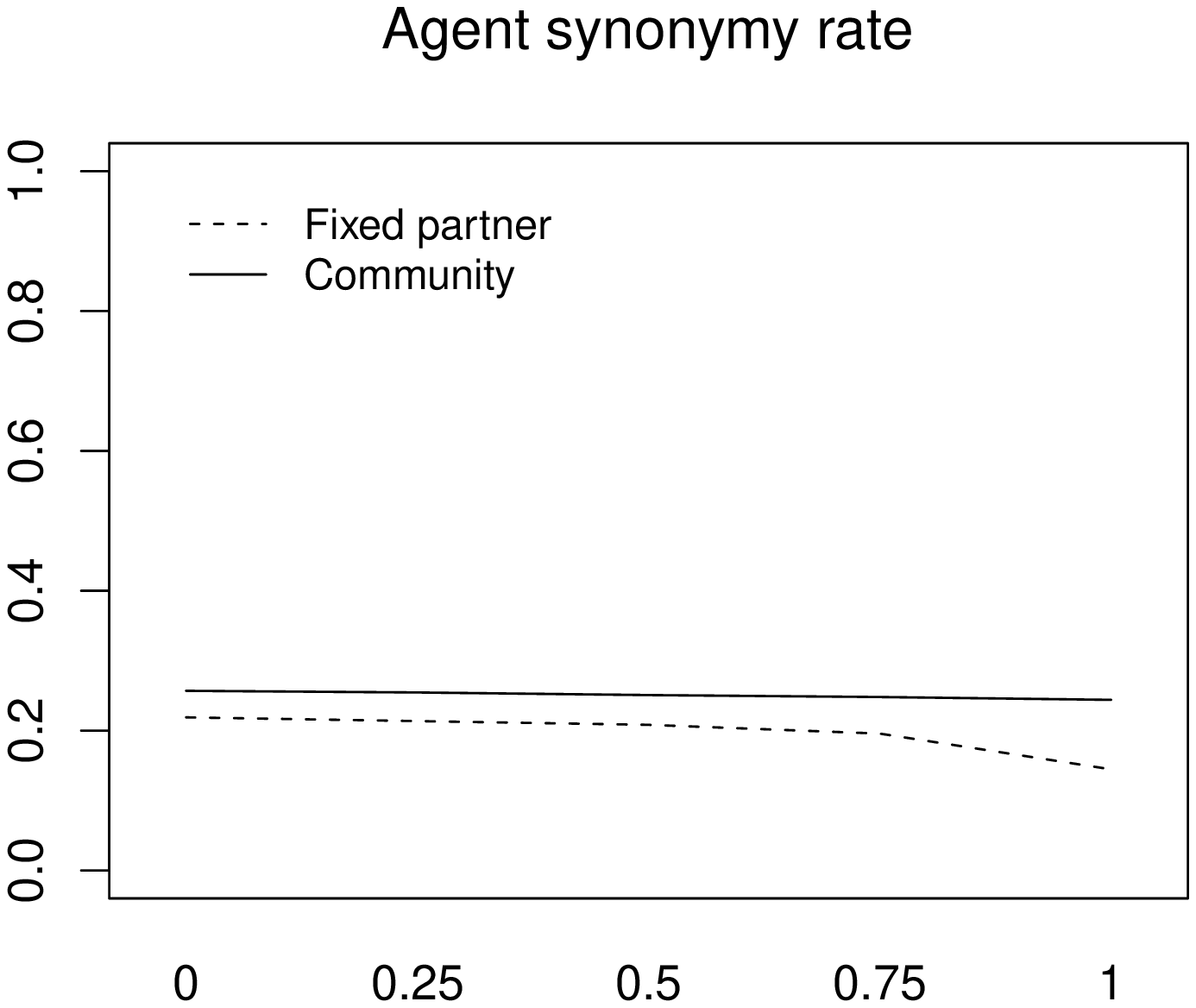}
\label{e1-agsyn}
}
\subfigure{
\includegraphics[angle=-90,scale=0.40]{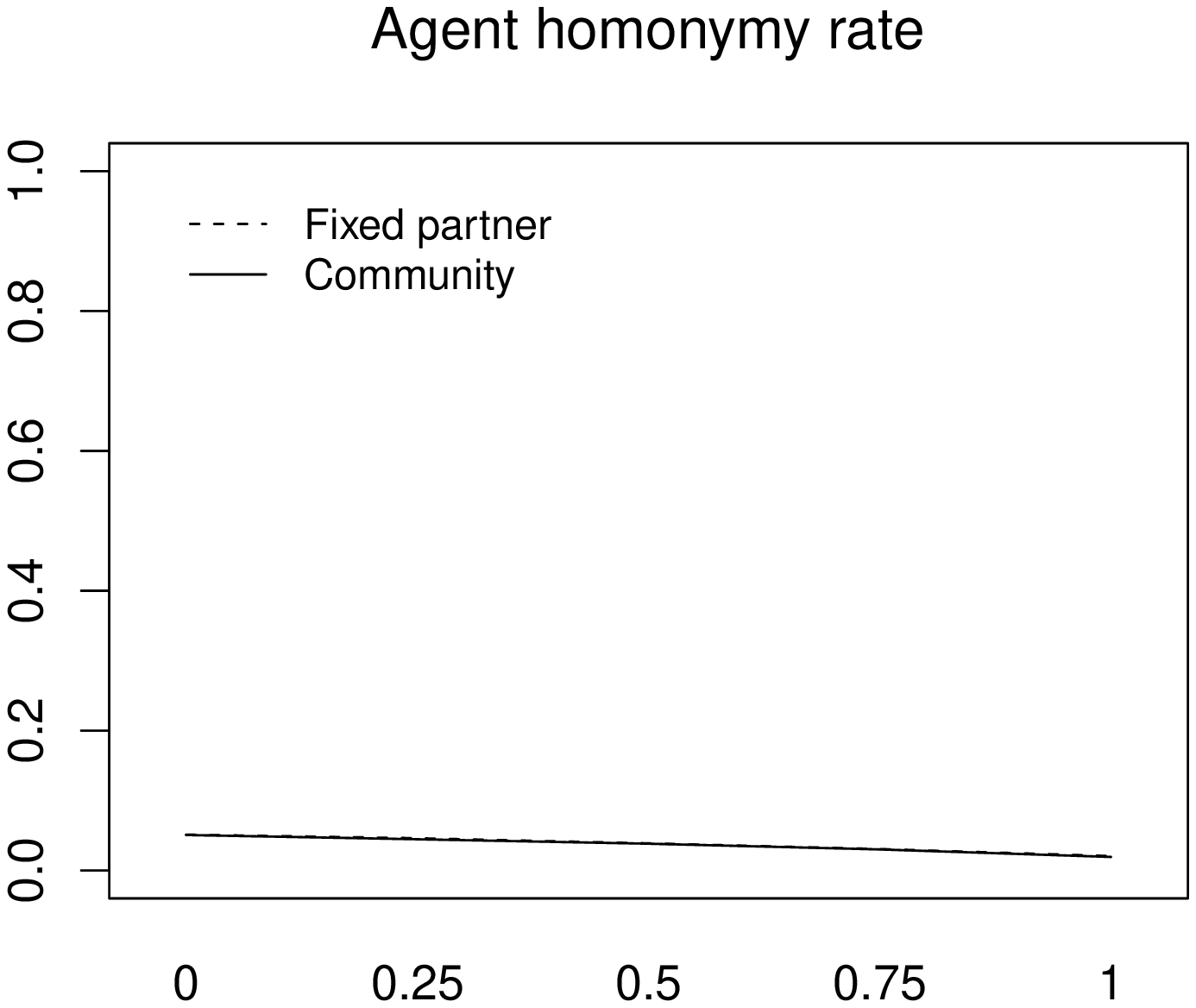}
\label{e1-aghom}
}
\caption{{\footnotesize Effect of the interaction type and the probability that success matters on \subref{e1-agsyn} agent lexicon synonymy and \subref{e1-aghom} agent lexicon homonymy.}}
\end{figure}

Figure \ref{e1-lbuse} depicts the rates of lexicon use, suggesting that with the increase of $p_{sm}$ and the corresponding diminishing of lexicon size ($t$ value between $-40.06$ and $-75.26$, $p<0.0001$), the number of \textit{forms} in an agent's lexicon appears to decrease ($t$ value between $-39.81$ and $-78.23$, $p<0.0001$) with a significant effect on lexicon use ($t$ value between $-4.57$ and $-20.38$, $p<0.0001$), as further confirmed by Figure \ref{e1-f}. The intuition is that for higher levels of $p_{sm}$, wrongly guessed meanings are not being recorded in the agents' lexicons, resulting in higher quality convention systems. This is confirmed by the increase in lexicon precision ($t$ value between $11.63$ and $101.64$, $p<0.0001$) depicted in Figure \ref{e1-lbprec}. Interestingly enough, the decrease in the number of different forms in agents' lexicons does not seem to have a significant effect on agent lexicon synonymy across the board ($p>0.1$ for $p_{sm}=0.25$; yet $t$ value between $-4.64$ and $-28.18$, $p<0.0001$ for higher levels of $p_{sm}$) (see Figure \ref{e1-agsyn}). Presumably, the reason for this is that the drop-off in the number of distinct meanings (see Figure \ref{e1-m}) is directly proportional to that of distinct forms, which would explain the less affected synonymy and homonymy ratios (see Figure \ref{e1-aghom} for a plot of the latter). 

\begin{figure}[htb]
\centering
\subfigure{
\includegraphics[angle=-90,scale=0.40]{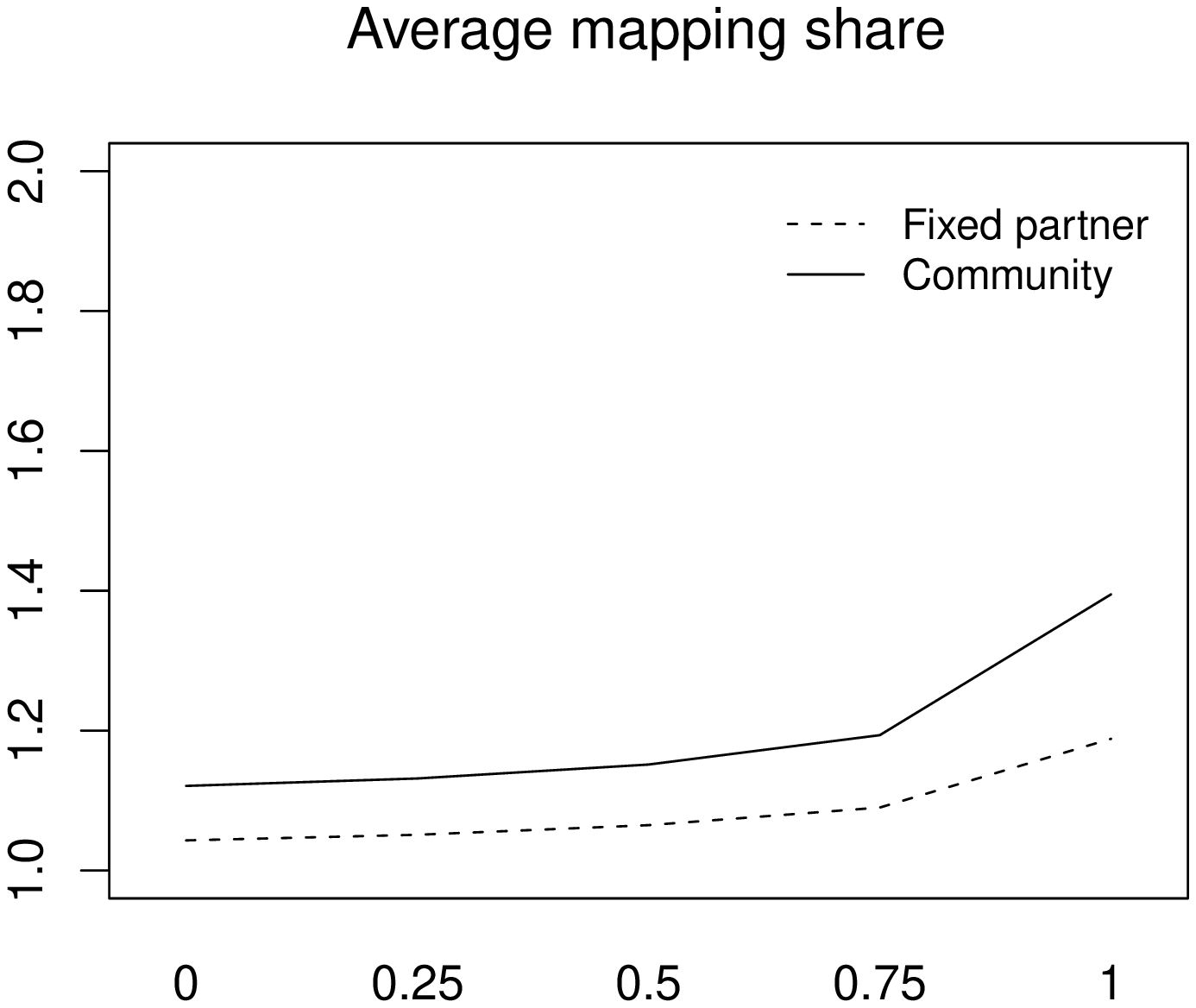}
\label{e1-mapshare}
}
\subfigure{
\includegraphics[angle=-90,scale=0.40]{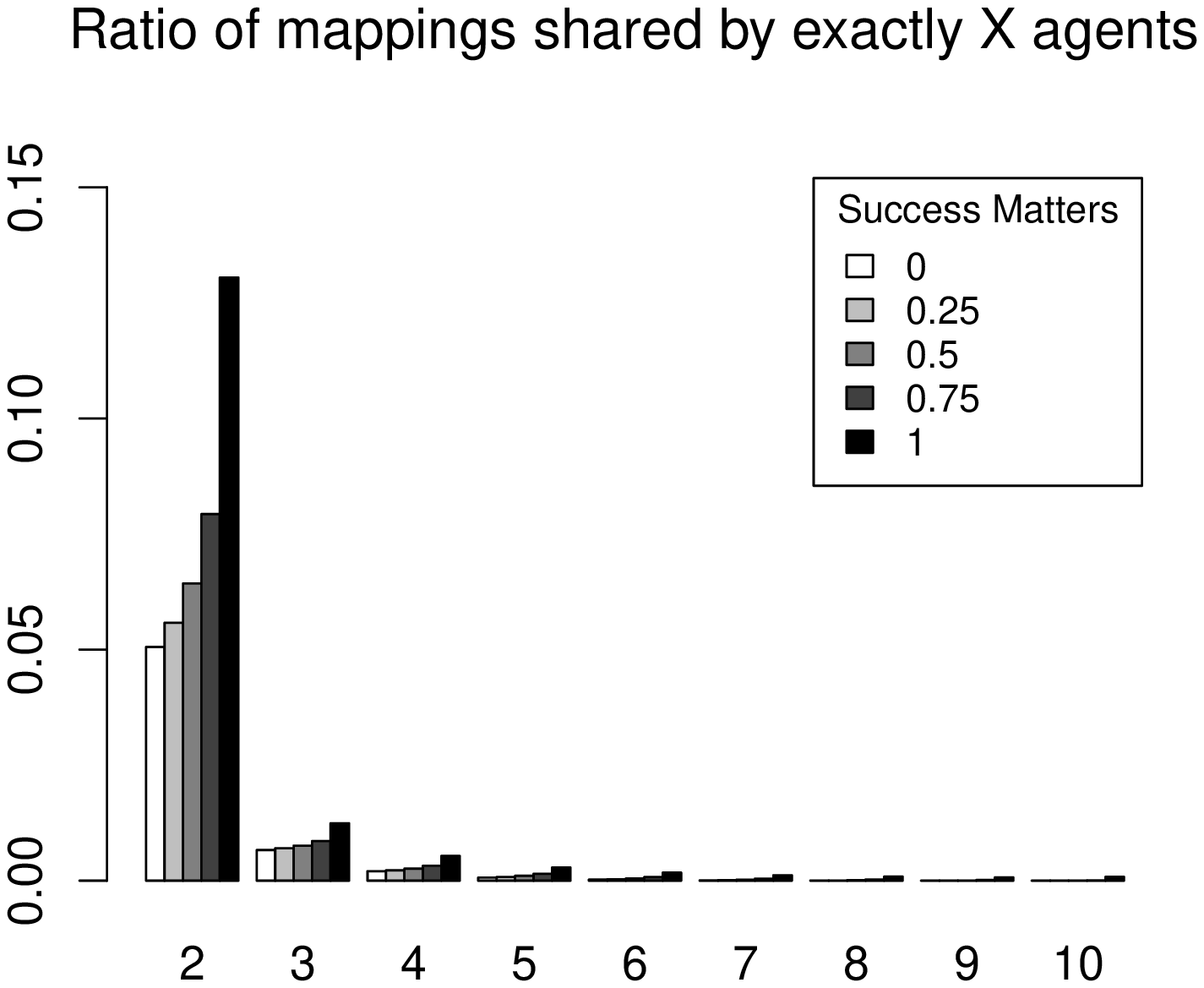}
\label{e1-mapshare-zoom}
}
\caption{{\footnotesize Effect of the interaction type (\subref{e1-mapshare} only) and the probability that success matters on \subref{e1-mapshare} average mapping share and \subref{e1-mapshare-zoom} ratio of mappings shared by exactly X agents.}}
\end{figure}

So far we have only evaluated the results of our experiments from the point of view of the agents, either by looking at their observed interaction success or by evaluating the communicative potential of their lexicons. However, as one of the main topics of the presented study was the establishment of conventions in a community of interlocutors, we should also evaluate the simulation results from the point of view of conventions, i.e. meaning-form mappings. In fact, there is a significant effect of both the community setting ($t$ value between $3.58$ and $91.14$, $p<0.00035$) and success mattering ($t=86.64$, $p<0.0001$) on the number of agents that share a mapping on average, as depicted in Figure \ref{e1-mapshare}. This effect is broken down in Figure \ref{e1-mapshare-zoom}, in which one can see the portion of the global lexicon that is shared by any particular number of agents.\footnote{The remainder of the mappings is not shared, i.e. known by only one agent.} The effects observed in the latter figure can be further described by an equation like $mapshare = a^{p_{sm}}*b^{-n}$, whereby the ratio of shared mappings ($mapshare$) is directly proportional to success mattering ($p_{sm}$) and inversely proportional to the number of agents ($n$) that are expected to know the mappings.

\section{Conclusions and Future Work}

In summary, experiencing a degree of success provides the all important foundation required for establishing linguistic conventions in task-oriented dialogue and dispersing these throughout the community. The ramifications of this finding are that language is very unlikely to have emerged for the benefit of a success-agnostic activity, such as gossip (cf. \cite{Dunbar1997}), but has presumably evolved as an adaptational necessity in times where human cooperation has become essential.

The shortcomings of the community setting can be attributed to the LEW's implementation of interactions as two autonomous activities and the lack of success-based adjustment of mapping usage strategies. Future work should aim to improve this aspect by looking into the interactive alignment model (cf. \cite{Pickering2004}).

\bibliographystyle{splncs03}
\bibliography{../../bachwerk}

\end{document}